# A Vision–Language–Action Model with Visual Prompt for OFF-Road Autonomous Driving

Liangdong Zhang[1], Yiming Nie[1*], Haoyang Li[2], Fanjie Kong[3], Baobao Zhang[1], Shunxin Huang[1], Kai Fu[1], Chen Min[4] and Liang Xiao[1]

[1] Defense Innovation Institute, Academy of Military Sciences, Beijing 100071, China
[2] Information Engineering College, Nanchang University, Nanchang 330031, China
[3] Computer Science College, Xi'an Jiaotong University, Xi'an 710072, China
[4] Institute of Computing Technology, Chinese Academy of Sciences, Beijing 100190, China
mincheng@ict.ac.cn

**Abstract.** Efficient trajectory planning in off-road terrains presents a formidable challenge for autonomous vehicles, often necessitating complex multi-step pipelines. However, traditional approaches exhibit limited adaptability in dynamic environments. To address these limitations, this paper proposes OFF-EMMA, a novel end-to-end multimodal framework designed to overcome the deficiencies of insufficient spatial perception and unstable reasoning in visual-language-action (VLA) models for off-road autonomous driving scenarios. The framework explicitly annotates input images through the design of a visual prompt block and introduces a chain-of-thought with self-consistency (COT-SC) reasoning strategy to enhance the accuracy and robustness of trajectory planning. The visual prompt block utilizes semantic segmentation masks as visual prompts, enhancing the spatial understanding ability of pre-trained visual-language models for complex terrains. The COT-SC strategy effectively mitigates the error impact of outliers on planning performance through a multi-path reasoning mechanism. Experimental results on the RELLIS-3D off-road dataset demonstrate that OFF-EMMA significantly outperforms existing methods, reducing the average L2 error of the Qwen backbone model by 13.3% and decreasing the failure rate from 16.52% to 6.56%.

## 1 Introduction

Trajectory prediction in off-road terrain is a crucial challenge for autonomous vehicles in expanding the boundaries of autonomous driving. The characteristics of off-road environments are that they lack structured roads and can include various natural landscapes, such as forests, deserts, and mountains. These terrains are unpredictable and dynamic, with obstacles such as uneven ground, foliage, and lack of clear or standardized road markings. In recent years, the widespread application of deep learning has driven significant progress in the urban autonomous driving systems. The traditional pipelines mainly adopt a modular architecture (perception-planning-decision pipeline), with its subsystems being built upon rule-based algorithms. This approach has technical advantages such as clear module interfaces and interpretable decision-making processes, but it suffers from a core deficiency of insufficient adaptability to off-road environments.

With the breakthrough in Multimodal-Large-Language-Model (MLLM) technology, the VLA framework has demonstrated innovative potential in the field of trajectory prediction. Taking the EMMA [1] system proposed by Waymo as an example, its knowledge transfer ability, obtained through pre-training on massive image-text data using a Vision-Language-Model (VLM), significantly enhances the scene reasoning capability and decision interpretability of autonomous driving systems. Current VLA models excel on structured urban datasets such as nuScenes [2], but face dual challenges in unstructured off-road environments: on the one hand, sparse prior knowledge of roads and the absence of lane lines and traffic topology information are prone to induce insufficient spatial perception in VLMs; on the other hand, reasoning mechanisms such as the Chain-of-Thought (CoT) employed by OpenEMMA [3] are prone to generating unstable trajectory prediction results under off-road environments.

However, existing research indicates that there is still a significant gap in the research on VLA methods in the field of off-road trajectory prediction, with immense potential for application. This underscores the urgent need for innovative methodologies that highlight the potential of the VLA framework—a paradigm embodied by our proposed model, OFF-EMMA, which leverages VLM with visual prompts to predict off-road trajectories. This model has been evaluated on the RELLIS-3D [4] dataset and achieved leading results, demonstrating reliable trajectory prediction capabilities. This work not only expands the application boundaries of VLA models but also provides a novel approach for trajectory prediction in off-road environments. Our main works are summarized as follows:

• We introduce an end-to-end VLA framework designed for trajectory prediction in off-road environments. This framework utilizes open-source pre-trained VLMs as backbone, augmented by visual prompt to further enhance its spatial perception capability.

• By seamlessly integrating the novel COT-SC reasoning strategy and visual prompts, effectively stabilize the reasoning results of the model.

• We conducted experiments on the RELLIS-3D dataset to evaluate the performance of OFF-EMMA in end-to-end trajectory planning. The average L2 error decreased by 13.3%, and the failure rate dropped from 16.52% to 6.56%, demonstrating its effectiveness and adaptability.

3## 2 Related Work

### 2.1 Large Vision Language Models

Large Language Models (LLMs) have demonstrated remarkable in-context and generalization capabilities through pre-training on massive internet corpora, particularly excelling in in-context learning, instruction following, and complex reasoning tasks. This capability stems from the rich knowledge representations acquired by the model during training and its cross-task generalization properties. Based on the successful paradigm of LLMs, researchers have further proposed Vision-Language Models (VLMs) that successfully extend the reasoning capabilities of language models to multimodal domains by projecting the output feature space of visual encoders into the linguistic token embedding space. Current representative works such as GPT-4 [5], Llama [6], and Qwen VL [7] have demonstrated excellent visual understanding and multimodal reasoning capabilities in open-domain tasks. However, these models are primarily trained on static two-dimensional images or videos, lacking sufficient spatial reasoning capabilities for dynamic three-dimensional driving scenarios. Additionally, they are more susceptible to hallucinations and overconfident misinterpretations when facing complex scenarios such as unstructured roads, which may lead to critical decision-making risks. These limitations severely restrict the application potential of VLMs in autonomous driving in off-road environments.

### 2.2 Vision-Language-Action model for Driving

With the breakthrough in multimodal alignment capabilities of VLMs, researchers have begun to explore the application of VLM/MLLM joint architectures in auto drive systems. This field has witnessed a series of advancements: LMDrive [8] pioneered the introduction of the CoT reasoning mechanism into the driving domain, constructing a human-vehicle natural language interaction system; the EMMA framework developed based on Gemini demonstrates the innovative potential of the VLA model by unifying the perception input and driving output through natural language, directly predicting vehicle trajectories, target objects, and road topology based on camera images, significantly enhancing motion planning performance. Subsequent research has continued to deepen this paradigm. OpenEMMA promotes technological transparency by integrating CoT reasoning technology and open-sourcing the complete model framework. DriveGPT4 [9], based on the Llama 2 architecture and combining the BDD-X dataset with pseudo-labeled data generated by ChatGPT, achieves collaborative optimization of multi-frame video temporal understanding and vehicle control prediction; DriveMLM [10] innovatively integrates structured driving rules, dynamic user intentions, and multimodal sensor data into the behavior planning layer, enhancing decision interpretability. Despite significant progress made by these models in urban environments, their training and evaluation still heavily rely on urban road datasets such as Waymo and nuScenes, lacking the ability to expand and generalize to off-road environments.

### 2.3 Visual Prompt

Visual prompts have been widely utilized for the transfer and adaptation of various downstream tasks, and can be categorized into learnable and image-modifying methods. The learnable visual prompt method incorporates trainable labels as additional visual inputs. Studies such as LM-BFF and VPT have demonstrated enhanced learning efficiency through prompt-based fine-tuning. The image-modifying visual prompt method focuses on altering images using expert-generated elements. Among them, FGVP [11], API [12], and SoM [13] have achieved significant improvements in visual understanding of VLMs through techniques such as segmentation masks and attention heatmaps. We drew inspiration from SoM, which applies semantic masks and label overlays onto images. We introduced a lightweight segmentation label block and embedded the meaning of the mask label into the prompt, significantly enhancing the spatial perception ability of VLMs.

## 3  Methodology

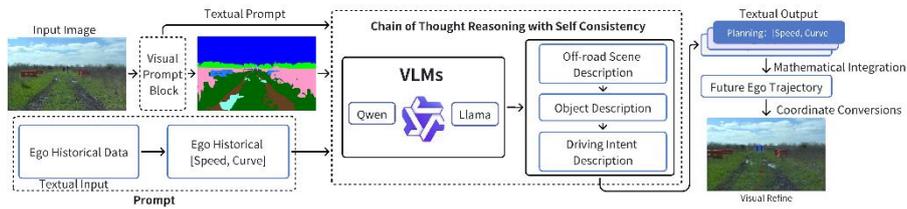

**Fig. 1.** Overall architecture of OFF-EMMA

Figure 1 illustrates the overall architecture of OFF-EMMA, an end-to-end off-road autonomous driving framework that utilizes open-source pre-trained VLMs to achieve trajectory planning and scene perception functions. It takes historical driving ego states and preprocessed images as inputs to predict future motion trajectories.

For each inference cycle, the front camera image from the RELLIS-3D dataset is first processed by the Visual Prompt Block (VP-Block) to obtain a labeled mask image, which is then input into the VLM along with the vehicle's historical ego driving data. To improve interpretability and prediction stability, we employ the COT-SC strategy. In the subsequent stage, it explicitly outputs a series of predicted future ego status. These prediction results, after numerical integration, generate the predicted trajectory, which is subsequently compared with the ground truth. Finally, after interpolation, the visualization results are obtained.

**3.1 COT-SC End-to-End Planning**

To achieve more efficient trajectory planning in complex off-road environments, we introduce the COT-SC strategy, enhancing the traditional approach based on CoT reasoning. COT-SC generates answers through multiple-path reasoning, effectively reducing the impact of outliers on the reasoning results, thereby improving the accuracy and robustness of predictions. Specifically, during the trajectory planning process, we first perform multiple-path reasoning to obtain several candidate results, then remove outliers and average the remaining predictions to obtain a more stable and realistic planning trajectory.

1) Reasoning Process: Firstly, we input the historical ego driving state (speed, curvature) and image into the VLM. Prior to this, the scene image is first processed by a semantic segmentation block to generate a mask as visual prompts, thereby providing the model with more accurate environmental semantic context. Subsequently, we designed a dedicated prompt for the VLM tailored for off-road tasks, which enables comprehensive analysis of the scene through multiple inferences in the COT. Specifically, it covers the following four aspects:

• Scene Description: Briefly describe the important elements in the off-road environments, such as terrain, road conditions, obstacles, etc.

• Object Description: Identify key elements that require attention in off-road environments, such as obstacles or the dynamics of other vehicles and pedestrians, and infer the impact of these elements on vehicle decision-making.

• Intention Description: Based on the current off-road environments, clearly indicate the expected behavior of the self-driving vehicle, such as whether it needs to cross obstacles, make turns, or slow down.

• Driving action prediction: By combining the chain-of-thought reasoning process with historical driving states, the VLM infers the vehicle's driving actions for the next 5 seconds, and outputs two intermediate representation sequences formatted as $[(v_1, c_1), \ldots, (v_n, c_n)]$. $v$ represents speed, with the unit being m/s, indicating the magnitude of the vehicle's forward speed, which simulates the degree of human accelerator pedal depression; $c$ represents curvature, with the unit being 1/m, indicating the magnitude of the vehicle's turning, which simulates the degree of human steering wheel rotation. We adopt this representation method rather than explicitly outputting predicted trajectory points since VLMs often struggle with reasoning over coordinate-based physical models.

2) Trajectory Prediction: We define the velocity sequence output by the VLM as $S = \{s_t\}$, representing the magnitude of the vehicle's speed; and the curvature sequence output as $C = \{c_t\}$, representing the curvature of the vehicle's turn. The specific calculation process for a single trajectory is as follows:

Heading angle update: At each step, we update the heading angle based on the product of velocity and curvature $\theta_t$:

$$\theta_{t+1} = \theta_t + \int_t^{t+1} c(\tau)s(\tau)d\tau \tag{1}$$

Velocity component calculation: Based on the updated heading angle, we can calculate the velocity components in the x and y directions:

$$v_x(t) = s_t \cos(\theta_t) \tag{2}$$

$$v_y(t) = s_t \sin(\theta_t) \tag{3}$$

**Prompt step1: Scene Description**
The vehicle is driving through a flat terrain with green grass and occasional water. The road is generally in good condition, though it shows some uneven patches. No significant obstacles or dynamic elements are present. The overall scene suggests a relatively smooth and open environment suitable for driving.

**Prompt step2: Object Description**
Road User 1: The pedestrian
- Location: The pedestrian is located near the center of the image, standing on the grassy area.
- Description: The pedestrian appears to be standing still, possibly observing something or waiting for someone.
- Importance: Pedestrians can pose a significant risk to drivers, especially if they suddenly step into the road or cross unexpectedly. It's crucial to maintain a safe distance from pedestrians and be prepared to stop if necessary.

**Prompt step3: Intent Description**
- The vehicle should maintain its current speed and continue straight. The terrain appears to be relatively flat with no significant obstacles or changes in elevation that would require steering adjustments. There are no dynamic objects present in the scene, so there is no need for the vehicle to change its speed.

**Prompt step4: Generate Motion**
- [0.3, 0.0], [0.3, 0.0], [0.3, 0.0], [0.3, 0.0], [0.3, 0.0]

**Fig. 2.** Output example of VLM in OFF-EMMA

Trajectory calculation: The final trajectory of the vehicle is obtained by numerically integrating the velocity components.

$$x_{t+1} = x_t + \int_t^{t+1} v_x(\tau)d\tau \tag{4}$$

$$y_{t+1} = y_t + \int_t^{t+1} v_y(\tau)d\tau \tag{5}$$

Eliminating outliers: We perform N independent trajectory inferences, and obtain a trajectory coordinate sequence with a length of T seconds after each inference using the formula above $\{(x_1^{(n)}, y_1^{(n)}) \ldots, (x_t^{(n)}, y_t^{(n)})\}$. For each time step t, calculate the sum $P_t^x = \{x_t^{(1)}, x_t^{(2)}, \ldots, x_t^{(N)}\}$ and standard deviation as follows (taking $P_t^x$ as an example):

$$\mu_t^x = \frac{1}{N}\sum_{n=1}^{N} x_t^{(n)} \tag{6}$$

$$\sigma_t^x = \sqrt{\frac{1}{N}\sum_{n=1}^{N}\left(x_t^{(n)} - \mu\right)^2} \tag{7}$$

Calculate the absolute difference between the value in the pair and the mean, respectively, and then check whether the difference $diff_i$ is less than or equal to $2\sigma_x + \epsilon$, $\epsilon$ is a small positive constant. If the condition is met, retain the prediction; otherwise, discard it.

$$diff_i = \left|x_t^{(i)} - \mu_t^x\right| \tag{8}$$

All eligible predictor values $x_t^{(1)}, x_t^{(2)}, \ldots, x_t^{(M)}$ are retained, and their mean is calculated. Similarly, following the same process, we ultimately obtain:

$$\hat{x}_t = \frac{1}{m}\sum_{i=1}^{m} x_t^{(i)} \tag{9}$$

$$\hat{y}_t = \frac{1}{m}\sum_{i=1}^{m} y_t^{(i)} \tag{10}$$

In this method, the application of the COT-SC strategy enhances the robustness of trajectory planning. During multiple inference processes, the system is able to identify and remove outliers, thereby reducing the impact of single inference errors on the final trajectory. Ultimately, the obtained trajectory is the average of multiple inference results, exhibiting higher stability and accuracy. This optimization strategy makes trajectory planning in off-road driving tasks more reliable, especially in complex and dynamically environments.

### 3.2 Visual Prompt Block

In this study, to enhance the trajectory planning and reasoning capabilities of pre-trained VLMs in complex off-road environments, we propose a visual prompt annotation module. Although existing VLMs have demonstrated excellent performance in language understanding and simple visual reasoning tasks, their limited spatial understanding ability makes it difficult to generate high-quality reasoning results when dealing with complex spatial reasoning problems. Especially in complex off-road driving scenarios, the model's spatial perception ability is insufficient, which seriously affects its performance in trajectory planning and decision generation.

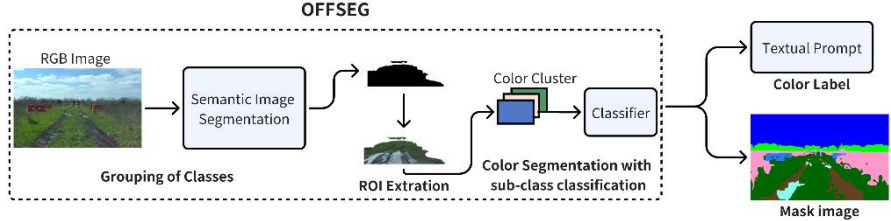

**Fig. 3.** VP-Block in OFF-EMMA

To address the limitations discussed above and enhance the performance of LLM in off-road tasks without requiring further fine-tuning of the pre-trained backbone, we were inspired by the MPDrive [14] and SoM methods and introduced an external open-source semantic segmentation model, OFFSEG [15], tailored for off-road tasks, which was then integrated into the OFF-EMMA system. This block performs efficient semantic segmentation on input RGB images and assigns unique color labels to different categories, thereby providing clear scene-level prompts for VLM.

OFFSEG works through a two-step pipeline, an overview of OFFSEG is represented in Figure 3. In the first step, the various labels are consolidated into four broad categories, and a semantic segmentation network is applied. In the second step, the resulting regions of interest are passed to a color-based segmentation module, which further divides them into finer subclasses—such as grass, mud, or puddles— and produces these detailed labels as the final output. In this way, the original image is converted into a scene feature map with clear spatial semantics, enabling the VLM to more accurately understand the traversability and obstacle information of different areas.

Furthermore, we integrate the color labels generated by the visual prompt block with traditional textual prompts, further enhancing the spatial perception ability of the VLM. Specifically, we introduce semantic descriptions for each color label when constructing the prompt, guiding the model in associating visual regions with semantic meanings during inference. By incorporating rich spatial semantic information, the VLM, which originally relied on language and simple visual input, is able to achieve deeper spatial reasoning, significantly improving its performance in complex off-road environments.

In summary, the introduction of the VP-Block enhances the spatial reasoning capability of the OFF-EMMA model in off-road scenarios. Especially when facing complex obstacles and dynamic environments, the model is able to make more precise and safer decisions based on spatial prompt information. This innovative approach not only significantly improves system performance but also avoids fine-tuning of the pre-trained model, greatly enhancing the model's adaptability and application flexibility.

# 4 Experiments

In this section, we conducted comprehensive end-to-end trajectory planning experiments in off-road environments, utilizing various VLMs to demonstrate the effectiveness and robustness of OFF-EMMA. Additionally, we performed ablation studies to delve into the contributions of each block within OFF-EMMA.

## 4.1 Dataset

The experiment was conducted on the RELLIS-3D off-road dataset, a purpose-built multimodal dataset for off-road environments. It encompasses various sensor data, such as lidar, cameras, and IMU, offering high-resolution imagery, dense LiDAR point clouds, and accurate pose information. The dataset features diverse off-road terrains, which are crucial for scene understanding and trajectory planning tasks. The dataset is divided into five scenes, with a total of 13,556 images.

## 4.2 Evaluation Metrics

We adopt the standard evaluation method for trajectory prediction tasks on nuScenes, defining Ground Truth (GT) as the actual vehicle trajectory of the vehicle. We calculate the L2 loss with GT at intervals of 1 second, 2 seconds, and 3 seconds, as well as their average value. Furthermore, when the average L2 loss exceeds 10m, the prediction is considered a failure.

## 4.3 Experimental Setup

We conducted our experiments on a server equipped with an RTX 3090 (24GB). The backbone models used in the experiments included Qwen2.5-VL-7B-Instruct and Llama-3.2-11B-Vision-Instruct. For fair comparison, we updated the Qwen2-VL-7B-Instruct backbone model used in OpenEMMA with the latest Qwen2.5-VL-7B-Instruct. For comparison, we employed a zero-shot method as a baseline, which relies solely on historical vehicle states and input images without involving any inference process. Additionally, we compared it with LightEmma [16], a lightweight version of OpenEmma utilizing COT. The results are presented in Table 1.

To further validate the robustness of OFF-EMMA in highly dynamic and unpredictable off-road environments, we conducted additional experiments focusing on three types of obstacle scenarios: static obstacles, dynamic obstacles, and suddenly emerging obstacles. These scenarios commonly occur in real-world off-road settings, where the lack of structured roads and prior knowledge leads to increased decision-making complexity. In particular, sudden obstacle emergence presents a major challenge to existing planning frameworks due to the absence of relevant historical perception. We selected 60 representative test cases from the RELLIS-3D dataset and constructed or augmented them as follows:

Static obstacle scenes were selected based on terrain images with trees, rocks, or fixed barriers directly in the vehicle's path.

Dynamic obstacle scenes were created by simulating object motion (e.g., moving vehicles or pedestrians) across a time sequence.

Sudden obstacle scenes were manually synthesized by inserting an obstacle into the last frame(s) of a previously obstacle-free sequence, simulating abrupt emergence.

Each scenario was evaluated using OFF-EMMA with Qwen2.5-VL-7B-Instruct backbone, as well as a zero-shot baseline. The success rate indicates whether the predicted trajectory successfully avoided the obstacle.

**Table 1.** Evaluation of Trajectory Prediction Performance on RELLIS-3D

| Method | Model | L2(m) 1s | L2(m) 2s | L2(m) 3s | L2(m) Avg | Failure Rate(%) |
|---|---|---|---|---|---|---|
| Zero-shot | Llama-3.2-11B-Vision-Instruct | 1.64 | 1.82 | 2.26 | 1.91 | 80.74 |
| | Qwen2.5-VL-7B-Instruct | 1.39 | 1.67 | 2.03 | 1.73 | 56.51 |
| OpenEMMA | Llama-3.2-11B-Vision-Instruct | 1.23 | 1.56 | 1.93 | 1.59 | 36.11 |
| | Qwen2.5-VL-7B-Instruct | 0.95 | 1.02 | 1.29 | 1.12 | 16.52 |
| LightEMMA | Llama-3.2-11B-Vision-Instruct | 1.26 | 1.61 | 1.96 | 1.67 | 35.93 |
| | Qwen2.5-VL-7B-Instruct | 0.93 | 1.07 | 1.33 | 1.18 | 17.62 |
| **Ours** | Llama-3.2-11B-Vision-Instruct | 1.17 | 1.48 | 1.77 | 1.41 | 20.76 |
| | **Qwen2.5-VL-7B-Instruct** | **0.88** | **0.95** | **1.06** | **0.97** | **6.56** |

**Table 2.** Obstacle Avoidance Performance Evaluation

| Model | Scene Type | Success Rate(%) | L2(m) Avg | Failure Rate(%) |
|---|---|---|---|---|
| OFF-EMMA | Static Obstacle | 75 | 0.99 | 10.6 |
| | Dynamic Obstacle | 65 | 1.13 | 13.7 |
| | Sudden Obstacle | 51.6 | 1.21 | 20.8 |
| Zero-shot | All (avg) | 31.6 | 2.23 | 36.7 |

### 4.4 Results and Discussion

Table 1 summarizes the performance evaluation results of OFF-EMMA across all scenarios in the RELLIS-3D dataset. The experimental results indicate that our method achieves overall performance improvements on both mainstream VLM backbone networks.

Specifically, the average L2 error of the Qwen version decreases from 1.12m to 0.97m, a reduction of 13.3%; the failure rate significantly decreases from 16.52% to 6.56%, validating the effectiveness of the CoT-SC strategy and the VP-Block in suppressing large trajectory deviations. It is worth noting that a performance gain of 11.3% is also achieved on the Llama-3-Vision backbone network, demonstrating the generalizability of the proposed method across different pre-trained backbones.

The prediction errors of all comparative methods exhibit an upward trend with cumulative time, but OFF-EMMA demonstrates the smallest error growth rate. We speculate that this is attributed to the dual advantages of the COT-SC mechanism: expanding the coverage of the solution space through diversified sampling and effectively reducing the systematic bias in velocity-curvature estimation by incorporating an outlier removal strategy.

As shown in Table 2, OFF-EMMA demonstrates superior performance across all obstacle types. In particular, the success rate in static obstacle scenarios exceeds 75%. These results confirm that the model's ability to react appropriately under abrupt environmental changes. These additional evaluations demonstrate that OFF-EMMA is capable of generating robust and adaptive trajectories even under sudden and dynamic obstacle interference. This robustness is especially critical for real-world off-road deployment, where environment uncertainty is a dominant challenge.

### 4.5 Visual Analysis

Figure 4 visualizes four representative decision-making cases of OFF-EMMA in typical off-road environments. Experimental results based on the Qwen backbone network demonstrate that:

In the straight-ahead scenario depicted in Figure 4a, the vehicle must maintain its course through a narrow passage where muddy ruts and puddles are interspersed.

In the right turn scenario depicted in Figure 4b, a smooth and feasible trajectory planning is demonstrated.

In the obstacle avoidance scenario depicted in Figure 4c, the timely detection of obstacles on the left side and the execution of emergency steering demonstrate the obstacle avoidance capability in complex terrains.

In the left-turn scenario depicted in Figure 4d, the available safety buffer on the left is effectively utilized to generate a stable trajectory.

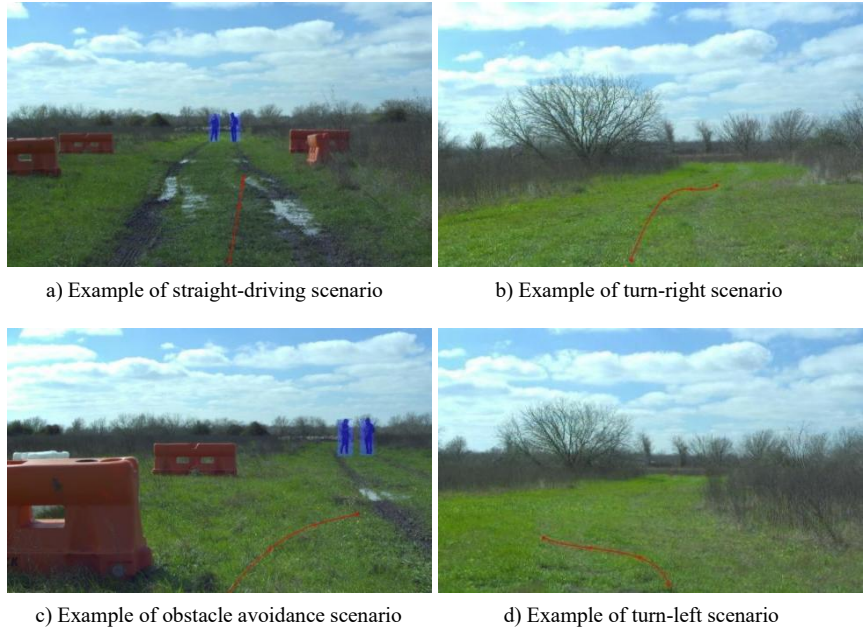

a) Example of straight-driving scenario    b) Example of turn-right scenario

c) Example of obstacle avoidance scenario    d) Example of turn-left scenario

**Fig. 4.** Example of visual cases

### 4.6 Ablation Experiments

The results in the table 3 show that the addition of VP-Block reduces the average L2 error from 1.12m to 1.02m, a decrease of 8.9%, and the failure rate significantly decreases from 16.52% to 7.82%. This confirms that incorporating visual prompts can enhance the spatial cognitive ability of VLMs, enabling them to achieve more accurate traversability reasoning and trajectory decision-making in terrains such as grasslands, bushes, and ruts. Comparative experiments further reveal that replacing COT-SC with the CoT strategy results in significant performance degradation, with an error increase of 10.4%. This phenomenon indicates that the self-consistency mechanism effectively suppresses occasional reasoning errors.

**Table 3.** Ablation experiments on RELLIS-3D

| VP-Block | COT-SC | L2(m) 1s | L2(m) 2s | L2(m) 3s | L2(m) avg | Failure rate(%) |
|---|---|---|---|---|---|---|
| - | - | 0.95 | 1.02 | 1.29 | 1.12 | 16.52 |
| ✓ | - | 0.90 | 0.99 | 1.17 | 1.02 | 7.82 |
| - | ✓ | 0.93 | 1.11 | 1.36 | 1.09 | 12.87 |
| ✓ | ✓ | **0.88** | **0.95** | **1.06** | **0.97** | **6.56** |

In summary, VP-Block provides explicit spatial semantics to enhance the spatial reasoning performance of VLMs; the COT-SC strategy reduces occasional errors through consistency evaluation, and removing either component individually results in performance degradation, indicating that both VP-Block and COT-SC are necessary.

## 5      Conclusion

This paper proposes an off-road end to end trajectory prediction model, OFF-EMMA, based on a VLA framework. Solving the issues of insufficient spatial reasoning capability and unstable decision-making in existing VLA models in unstructured environments. Experimental results demonstrate that OFF-EMMA achieves leading performance on the RELLIS-3D dataset, with an average L2 trajectory error and failure rate reduced by 13.3% and 60.3% respectively compared to the baseline method, and exhibits good generalization across various backbone networks. Furthermore, OFF-EMMA exhibits effectiveness and robustness in challenging off-road driving scenarios, providing a more effective solution for trajectory planning tasks in off-road environments. This work highlights the great potential of VLMs in the field of off-road trajectory prediction. In future work, we plan to explore multi-sensor fusion mechanisms to further enhance the spatial perception and planning capabilities of VLMs in off-road environments.